%% file: DataAnnotation.tex
\documentclass[10pt,twocolumn,letterpaper]{article}

\usepackage{iccv}
\usepackage{times}
\usepackage{epsfig}
\usepackage{graphicx}
\usepackage{amsmath}
\usepackage{amssymb}
\usepackage{times}
\usepackage{soul}
\usepackage{url}
\usepackage[utf8]{inputenc}
\usepackage[small]{caption}
\usepackage{booktabs}
\usepackage{diagbox}
\usepackage[flushleft]{threeparttable}
\usepackage{multirow}
\usepackage[accsupp]{axessibility} 

\usepackage[marginal]{footmisc}

\usepackage[pagebackref=true,breaklinks=true,letterpaper=true,colorlinks,bookmarks=false]{hyperref}

\iccvfinalcopy 

\ificcvfinal\fi

\input{ourCommands}

\begin{document}

\title{RaidaR: A Rich Annotated Image Dataset of Rainy Street Scenes}

\author{Jiongchao Jin \inst{1,2} \thanks{}
\and Arezou Fatemi \inst{1} \footnotemark[1]
\and
Wallace Michel Pinto Lira\inst{1}
\and
Fenggen Yu\inst{1}
\and 
Biao Leng\inst{2}
\and
Rui Ma\inst{3,1} \thanks{} 
\and
Ali Mahdavi-Amiri\inst{1}
\and 
Hao Zhang  \inst{1}
}

\institute{
\inst{1}Simon Fraser University \and
\inst{2}Beihang University \and 
\inst{3}Jilin University \and 
}

\twocolumn[{%
\renewcommand\twocolumn[1][]{#1}%
\maketitle
\begin{center}
    \includegraphics[width=1\textwidth]{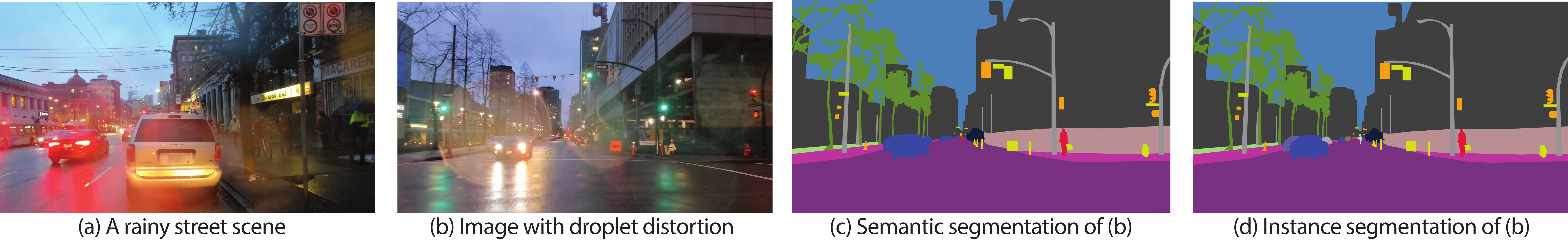}
    \captionof{figure}{RaidaR is a new annotated dataset to support autonomous driving research. It contains the {\em largest\/} number of {\em rainy street scene images\/} (58,542) to date --- 5,000 of them provide semantic segmentations and 3,658 with object instance segmentations.}
    \label{fig:teaser}
\end{center}%
}]
\ificcvfinal\thispagestyle{empty}\fi

\footnote{\quad $^{*}$Jiongchao Jin and Arezou Fatemi contribute equally to this work.}
\footnote{\quad $^{\dagger}$Rui Ma is the corresponding author.}


\input{MainAbstract}
\input{MainIntro}
\input{MainRW}

\input{Maindataset}
\input{Mainexperiments}

\input{MainMasked}

\input{MainConclusion}

{\small
\bibliographystyle{ieee_fullname}
\bibliography{egbib}
}

\newpage
\input{DataAnnotationSupp_arXiv_Appendix}

\end{document}

%% file: ourcommands.tex
\usepackage{color}
\definecolor{turquoise}{cmyk}{0.65,0,0.1,0.1}
\definecolor{purple}{rgb}{0.65,0,0.65}
\definecolor{dark_green}{rgb}{0, 0.4, 0}
\definecolor{dark_blue}{rgb}{0, 0, 0.4}
\definecolor{orange}{rgb}{0.6, 0.3, 0.0}
\definecolor{red}{rgb}{0.8, 0.2, 0.2}
\definecolor{light_red}{rgb}{0.8, 0.5, 0.5}

\newcommand{\am}[1]{{\color{black}#1}}
\newcommand{\wl}[1]{{\color{black}#1}}

\newcommand{\rui}[1]{{\color{black}#1}}
\newcommand{\Lagr}{\mathop{\mathcal{L}}}

%% file: MainAbstract.tex
\begin{abstract}
We introduce RaidaR, a rich annotated image dataset of rainy street scenes, to support autonomous driving research. The
new dataset contains the {\em largest\/} number of rainy images (58,542) to date, 5,000 of which provide semantic segmentations 
and 3,658 provide object instance segmentations. 
The RaidaR images cover a wide range of realistic rain-induced artifacts, including fog, droplets, and road reflections,
which can effectively augment existing street scene datasets to improve data-driven machine perception during rainy 
weather.
To facilitate efficient annotation of a large volume of images, we develop a semi-automatic scheme combining 
manual segmentation and an automated processing akin to cross validation, resulting in 10-20 fold reduction on annotation time.
We demonstrate the utility of our new dataset by showing how data augmentation with RaidaR can elevate the accuracy of 
existing segmentation algorithms.
We also present a novel unpaired image-to-image translation algorithm for adding/removing rain artifacts, which directly 
benefits from RaidaR.
%
\end{abstract}

%% file: MainIntro.tex
\section{Introduction}
\label{sec:intro}

Autonomous driving under adversarial weather conditions is challenging due to sensors introducing image artifacts and quality degradation~\cite{sensorDegradation}
that consequently hinder the performance of visual perception algorithms~\cite{Steinhauser2021,Mehra2020,Ort2020,Blin2019,zhou2019}.
The most prominent such weather conditions
is rain. Images captured during rainy weather can be blurry, with partial coverage and non-linear distortion caused by rain 
streaks or droplets; see Figure~\ref{fig:teaser}(b). 
Even before or after the rain, gloomy lighting conditions can create difficulties for color perception and 
delineation, and intensive use of vehicle headlights and street reflections due to water accumulation near 
night times, as shown in Figure~\ref{fig:teaser}(a), are problematic as well.
Machine perception tasks critical to autonomous driving such as semantic and instance segmentation~\cite{zhou2019,treml2016speeding,blum2019fishyscapes,chen2017importance,kirillov2019panoptic,zhou2020joint,neven2017fast,wong2020identifying,homayounfar2020levelset,mohamed2020instancemotseg} are most negatively impacted by
these artifacts from rainy scenes.


In this paper, we introduce RaidaR, a new dataset that is {\em rich\/} in providing street scene images under rainy weather, and it comes with {\em annotations\/} in the form of both {\em semantic\/} and object {\em instance segmentations\/}; see Figures~\ref{fig:teaser}(c-d). Our goal is to provide sufficient and diverse rainy scene data to augment existing datasets to improve the performance of data-driven autonomous driving algorithms. Specifically, RaidaR represents
the {\em largest\/} such dataset to date, consisting of 58,542 rainy images, with 5,000 of them annotated with semantic segmentation and 3,658 with instance segmentation. In addition, 4,085 sunny images were also annotated with semantic segmentations.

\input{tableRW}

To acquire RaidaR, we employed a roof-mounted camera platform to collect RGB videos under sunny and rainy weather conditions in Metro Vancouver, Canada. Among a total of 77,437 frames extracted from the collected videos, there are 18,895 sunny and 58,542 rainy images on which human faces and license plates had been identified and blurred to respect privacy. The captured rainy images cover various characteristics of rainy weather conditions including fog, road reflection, water droplets, etc.
Figure \ref{fig:teaser} shows some sample images from RaidaR.



In Table~\ref{tab:datasets}, \am{we list a set of known datasets.
Our main goal is to include datasets providing real (not synthetic) images with rainy scenes accompanied with GT segmentations. For reference, we also added the best known large-scale datasets including KITTI, Cityscapes, and Apolloscape.}
We focus on reporting stats related to rainy images and their annotations.
As we can see, the BDD dataset~\cite{BDD}, the largest in terms of total number of images, has the closest number of
rainy images compared to RaidaR, but RaidaR has 10 times as many. Only 253 rainy images in BDD have full-frame semantic segmentation, while
RaidaR offers thousands.
On the other hand, Raincouver~\cite{raincouver} was devoted to rainy weather, but it is quite small (326 images in total)
and its semantic segmentations only emcompass four object categories. In contrast, RaidaR offers 5,000 
semantically segmented rainy images on 19 object categories.

Another feature of our new dataset is that more than half (about 35K) of the rainy images have varying degrees of
image distortion caused by the presence of water droplets on the camera lens. Data-driven de-raining and droplet removal have 
been studied by several recent works~\cite{qian2018,porav019,porav2020rainy,MPR_net}, where they mainly work with
synthesized images composed using small droplets that were either produced by spraying water on a transparent shield in front of 
the camera lens or by synthesizing droplets considering their physical properties. In contrast, droplet-tempered images
in RaidaR were acquired under natural conditions mimicking realistic autonomous driving scenarios. Furthermore, a 
significant number of these images have been annotated. Algorithms trained on such data should generalize much better on real test data.

To facilitate efficient annotation of a large volume of images in RaidaR, we develop a semi-automatic technique combining
manual segmentation and an automated processing akin to {\em cross validation\/}. Specifically, we merge the results from 
four state-of-the-art segmentation networks with different weights to assess the reliability of each pixel label, where manual
intervention is called only when there is a large inconsistency. On average, this approach results in a 10-20 fold speedup over
completely manual segmentation.

Our paper makes the following contributions:
\begin{itemize}
\vspace{-2pt}
  \item We introduce the largest dataset of rainy street scene images to date, to support research in self-driving. The images cover a wide range of realistic rain-induced artifacts, including fog, droplets, and road reflections. The dataset contains thousands of rainy images with semantic and instance segmentations. 
\vspace{-5pt}
  \item We develop an efficient semi-automatic scheme for segmentation annotation, based on cross validation, to reduce manual labeling efforts.
\vspace{-5pt}
  \item We demonstrate the utility of our new dataset by showing how data augmentation with RaidaR can elevate the accuracy of existing segmentation algorithms.
\vspace{-5pt}
  \item We present a novel unpaired image-to-image translation algorithm, based on a {\em masked\/} version of GANHopper~\cite{lira_eccv20}, for adding/removing rain artifacts, which directly benefits from our annotated dataset.
\end{itemize}

%% file: tableRW.tex
\setlength\tabcolsep{3pt} 
\begin{table}
  \centering
      \small
  \begin{threeparttable}
  \begin{tabular}{l|r r r r}
\textbf{ Name}             & \textbf{\#Lab/\#Total} & \textbf{\#Rainy} & \textbf{\#SSeg} & \textbf{\#ISeg}      
\\ \hline
Raincouver~\cite{raincouver}                & 326/54K           & 326            & 326                       & 0                 \\ 
BDD~\cite{BDD}                          & 10K/120M         & 5.8K         & 253 & 253       \\ 
nuImages~\cite{caesar2020nuscenes}\tnote{*}   & 1.9K/18K          & 1.3K           & 58       & 58                                   \\ 
ACDC~\cite{acdc2021}             & 4K/4K        &    1K              & 1K       & 0 \\ 
Cityscapes~\cite{cityscapes}                & 5K/20K            & 0               & 0         & 0                                 \\ 
KITTI~\cite{Geiger2012CVPR}                       & 400/15K            &  0               & 0        & 0                                  \\ 
Apolloscape~\cite{huang2019apolloscape}             & 147K/147K        &  0                & 0       & 0 \\ 
RID~\cite{li2019single}             & 2.5K/2.5K        &    2.5K              & 0       & 0 \\ 
RaidaR                      & 9K/77K          & \bf 58,542  & \bf 5,000 & \bf 3,658                                        \\ 
\end{tabular}
  \end{threeparttable}
  \vspace{-5pt}
  \caption{Comparing the availability of rainy samples, and their annotations, from well-known street scene image datasets to RaidaR.
\#Lab/\#Total is the ratio between the number of labelled images (rain or no rain) and the total number of images in the datasets; \#Rainy is the number of rainy images; \#SSeg is the number of {\em rainy\/} images with semantic segmentations; and \#ISeg is the number of rainy images with instance segmentations.
  * Only front camera frames are considered. Weather labels were not originally available, the values shown here are manually curated.}    
    \label{tab:datasets}
    \vspace{-12pt}
\end{table}

%% file: MainRW.tex
\section{Related Work and Datasets}
\label{sec:RW}
Here we discuss road datasets for training data-driven approaches for autonomous vehicles and emphasize their differences in terms of weathering artifacts (e.g., rain) and semantic/instance segmentation. We then provide a brief review of road segmentation tools and techniques.

\subsection{Road Datasets}
\label{sec:datasetRW}
\am{There are many available datasets 
to train autonomous driving systems \cite{cityscapes,BDD,woodscape_dataset,darkzurich_dataset,acdc2021,huang2019apolloscape}.
Methods capable of synthesizing weather artifacts have been proposed along with efforts to capture natural scenes by driving on the streets at different times and weather conditions.}

Datasets to train autonomous cars may also carry various modes and information about the roads.
For instance, they may be composed of imagery datasets to capture visual contents on the roads, along with Lidar datasets that are rich sources of depth and geometry information~\cite{robotcar_dataset,utbm_dataset,choi2018kaist}.
\wl{Here, we provide an overview of different synthetic and real road datasets with semantic and instance segmentation.}

Synthesizing road datasets can be very useful since they are flexible in producing desired scenes that are difficult to capture by driving on the streets.
Frameworks such as CARLA \cite{Dosovitskiy17} have been designed to generate road data for training autonomous cars.
\wl{SYNTHIA~\cite{SYNTHIA_dataset}, made via Unity, provides approximately 200K samples, while GTA5~\cite{playing_dataset} uses footage from the Grand Theft Auto 5 computer game to generate 25K such frames.
Even though data synthesis is a great way to obtain a large amount of reliable segmented data, it has some limitations as they often fail to capture the complex nature of weather and road-specific phenomena. This may lead to training networks that are not robust in real-world applications. For instance, these works may fail to reproduce realistic droplets on the camera/windshield or road reflections.
}
Another way for simulating raining effects is to spray water on some transparent shield placed in front of the camera \cite{porav019, porav2020rainy}.
However, data captured in this way is still unlike real-world rainy images which contain overcast clouds, mist, and road reflections. 
Therefore, some efforts have been devoted to collect datasets under various weather conditions such as
Canadian Adverse Driving Conditions Dataset \cite{pitropov2020canadian}, Waymo open dataset \cite{sun2019scalability}, Lyft \cite{lyft_dataset}, A* 3D \cite{pham20203d}, Nuscenes \cite{caesar2020nuscenes}, etc.
However, there is no semantic segmentation ground truth accompanied with these datasets.
The Rain in Driving (RID) dataset \cite{li2019single} provides 2,496 real rainy images annotated with bounding boxes for selected traffic objects such as car, person, and bicycle. Hence, it can only be used for evaluating detection instead of segmentation algorithms.

Datasets with semantic segmentation ground truths exist but the size of rainy images in these datasets is not considerable. 
One of the earliest attempts to collect weathered datasets is Raincouver \cite{raincouver} which provides 30 minutes of video recorded in Vancouver in different weather conditions (mostly rainy).
This dataset, however, only provides 326 annotated images in which only four classes of objects are annotated (road, person, vehicle, and other).
While this is a great attempt, Raincouver does not include all the categories necessary for training an autonomous car such as traffic lights, sidewalks, etc.
In CamVid \cite{camvid}, only about 700 images are manually annotated with per-pixel semantic labels.
Cityscapes \cite{cityscapes}, which is the most well-known road dataset, contains 20K images with coarse and 5K images with per-pixel semantic annotation.
However, most samples from Cityscapes are captured under normal weather conditions with no rain artifacts.
\wl{However, there are weather-based augmentation techniques that have been applied on Cityscapes dataset, namely Weather Cityscapes \cite{weathercityscapes_dataset}, and Foggy Cityscapes \cite{foggyZurich_dataset}, that employ synthetic rain or fog effects. The main advantages of using such approaches are the extendibility of the method to other similar datasets and the possibility of reusing the expensive pixel level semantic and instance segmentation annotations for generating multi-modal samples.}

BDD \cite{BDD} is one of the most complete road datasets which contains more than 1,100 hours of videos. 
However, among 10K images with segmentation ground truth, only 253 of them have raining artifacts.
Appolloscape \cite{huang2019apolloscape}, another great effort that provides various data forms such as semantic point cloud, per-pixel semantics, etc. has considered collecting scenes with challenging weather conditions (such as rainy) as a future work.
\wl{ WildDash~\cite{wilddash_dataset} provides 4K of diverse images and their semantic segmentation ground truth, but this dataset is meant to be used as a benchmark for panoptic, semantic, and instance segmentation. They have not provided explicit weather classification of their dataset but after manually curating the dataset, we determined that it contains only 785 rainy scenes.
}
\rui{The recent ACDC dataset \cite{acdc2021} provides 4,006 images with high-quality semantic segmentation and the images are evenly distributed between fog, nighttime, rain, and snow.
Although ACDC is valuable for training and testing semantic segmentation methods on adverse visual conditions, it does not provide as many rainy images as RaidaR, nor the instance segmentation on images.
}

Therefore, despite the existence of real (non-synthetic) semantically segmented datasets, they are not \wl{as} rich in terms of rainy images \wl{as RaidaR}. 
Consequently, networks trained mainly on normal weather conditions perform poorly on rainy scenes.
Table \ref{tab:datasets} summarizes some datasets with ground truth segmentation and compares them with RaidaR.
As apparent in the table, RaidaR has the highest number of rainy images that are accompanied by ground truth semantic segmentation.
In Section \ref{sec:cross-testing}, we show that adding RaidaR for training segmentation networks will significantly improve the results.

\subsection{Segmentation Techniques}
\label{sec:segRW}
Semantic and instance segmentation are key algorithms when a vision system deals with inference (e.g., self-driving cars \cite{janai2020}).
Great efforts have been devoted to segment road scenes using deep neural networks \cite{minaee2020,janai2020,vpred}. 
Most of these networks are supervised therefore various tools have been proposed to help annotators indicate the right label of a pixel \cite{camvid,hitachi}.
However, manually labeling all the pixels is tedious and time consuming. 

Semi-automatic techniques have been introduced to ease and speed up this process. 
Mimicking the process of manual annotation, a polygon prediction is introduced in \cite{castrejon2017} where the polygon vertices are identified and manually adjusted on a cropped bounding box around a desired object.
This work is later improved by designing a new CNN encoder architecture and employing reinforcement learning and graph neural networks to elevate the accuracy and resolution of the results \cite{acuna2018}.
Curve-GCN \cite{ling2019} is the next attempt to improve these techniques by propagating messages via a graph convolutional network (GCN) to predict location adjustments of nodes in an iterative process.
However, in these methods, an initial bounding box is needed for each object and the final results need to undergo a manual adjustment of the polygon's vertices to capture fine details (see Section \ref{sec:experiment}).

With similar motivation to reduce the load of manual annotation, Vpred \cite{vpred} exploits video prediction models' ability to predict future labels. A joint propagation strategy is also employed to reduce misalignment in synthesized segmented images. 
Vpred produces high-quality results (close to ground truth) for many images especially when the image is not contaminated with weathering artifacts. 
However, it is unable to produce reasonable results on rainy scenes when it is trained on datasets with limited rainy instances.
To produce our ground truth, we use Vpred as one of the main deciders on the label of each pixel as it is one of the most reliable road segmentation networks.
We also cross validate its results against three other recent networks to make sure that a produced label is accurate (see Section \ref{sec:dataset}). 
For accurately separating instance segments or when these four networks are not unanimous, we ask for manual interventions. 
We show that this method significantly reduces the labor work involved in segmentation and produces close to ground truth results (Section \ref{sec:experiment}).

%% file: MainDataset.tex
\section{RaidaR Dataset}
\label{sec:dataset}

\subsection{Collection and Cleaning}
\label{sec:cleaning}
\vspace{-2pt} 
Our dataset is composed of various road types including mountains with natural scenery, downtown with traffics, high-ways, etc. We have also tried to collect images for the same routes under both sunny and rainy weather conditions to encourage consistency.
We mounted our camera on the roof of a sedan car with a small box around it to stop rain droplets blocking the lens too quickly. The employed camera was Logitech Ultra HD Webcam with 90 degree extended view recording 30 frames per second. After recording, we manually removed rainy images heavily blocked by droplets and sunny images with massive over-exposures.
After this step, our resulting dataset contains 34,951 rainy images with small droplets, 23,591 rainy images without droplets and 18,895 sunny images. The resolution of each image is 1920*1080.
To respect the privacy of people and vehicles appearing in our collected data, we detected human faces~\cite{li2019dsfd} and license plates~\cite{silva2018a} and performed a normalized box filter to blur them.

\subsection{Ground Truth Generation}
\label{sec:GTGEN}
\vspace{-2pt} 
To generate our ground truth, we perform a \emph{cross validation} mechanism to obtain the correct labels for each pixel. 
Specifically, we segment each image with four state-of-the-art segmentation networks and combine the segmentation labels of those networks to identify areas for which we were not able to find reliable labels.
The uncertain regions are then submitted to a \emph{manual intervention} step to annotate their right labels.
Finally, all other automatically labeled areas are inspected to confirm their accuracy.
This process significantly accelerates the ground truth labeling and reduces the manual work. 
Based on our labeling process experiment in Section \ref{sec:experiment}, it is 10 to 20 times faster to use our cross-validation based process than directly labeling all the pixels on original images. 

\vspace{4pt} 
\noindent\textbf{Cross Validation.} To automatically generate ground truth for semantic segmentation on RaidaR, we use four segmentation networks: Vpred \cite{vpred}, Accel \cite{jain2019accel}, PSP \cite{PSP} and PSA \cite{PSA} to generate the initial semantic masks. 
As different methods may predict different labels for each pixel, we cross validate the labels produced by each method by combining them with different weights and generating a \emph{confidence score} for all candidate labels of each pixel.
Formally, for each semantic mask with label $\ell$, we calculate the weighted value $S_{i,j}^\ell$ for each pixel $(i,j)$ as the confidence score of label $\ell$:\\ 
\begin{equation}
\label{eq:conf_score}
S_{i,j}^\ell = \sum_{k\ in\ \Pi} w_{k}M_{k}^\ell(i,j).
\end{equation}
Here, $\Pi=\{\text{Vpred, Accel, PSP, PSA}\}$ and $w_k$ is the weight of $k$th segmentation method; $M_{k}^\ell(i,j)$ is per-pixel binary mask value 
(1 if label $\ell$ has been predicted for pixel $(i,j)$ by method $k$ and 0 otherwise).

We assign the label with the largest confidence score to each pixel.
As different $w_k$ configurations may lead to different results, we discuss how to set proper $w_k$ in Section \ref{sec:Cross_Validation}.
We treat pixels with a confidence score higher than a threshold $\alpha$ as \emph{reliable} and others are treated as unreliable since they received contradictory labels from the networks.
We set a relatively high threshold $\alpha=0.7$ to ensure the correctness of cross-validated labels and we highlight the unreliable pixels for manual intervention.

\vspace{4pt} 
\noindent\textbf{Manual Intervention.}
After finding the pixels that need manual intervention, \am{our in-house team annotated the labels.} 
To ease the data annotation process, we offered the masks with regions of uncertain pixels and made a simple annotation tool. Using this strategy, annotators only needed to manually label $\approx24.8\%$ of an image.
Moreover, we asked users to inspect the labels that are automatically generated by the cross validation and correct them in case of false predictions.
Each person needs to spend 5-10 minutes to fully label one image depending on the size and complexity of the unreliable regions.
Based on our experiments, manually labeling an entire image takes 45-90 minutes.

\vspace{4pt} 
\noindent\textbf{Instance Annotation.}
Following the format of BDD and Cityscapes, we provide a distinct segment for every vehicle and pedestrian as they have higher priorities in autonomous driving. To produce instance segmentation masks, we used our ground truth semantic segmentation and assigned different colors to detached pedestrian and vehicle segments. \am{Then, our in-house team manually distinguished instances with overlaps.}

\begin{table*}[ht]
    \centering
    \small
    \setlength{\tabcolsep}{1.2mm}
    \begin{tabular}{c|c|c|c|c||c|c|c|c}
\multirow{2}{*}{\diagbox{Train}{Test}} & CS & BDD &BDD rainy& RaidaR & CS & BDD & BDD rainy & RaidaR \\
& \cite{vpred}/\cite{hmsa} & \cite{vpred}/\cite{hmsa} & \cite{vpred}/\cite{hmsa} & \cite{vpred}/\cite{hmsa} & \cite{pointrend}/\cite{tensormask} &\cite{pointrend}/\cite{tensormask}&\cite{pointrend}/\cite{tensormask}&\cite{pointrend}/\cite{tensormask}\\\hline
         RaidaR & 0.582/0.603 & 0.519/0.612 & 0.564/0.612 & 0.715/0.736
         & 0.318/0.324 & 0.303/0.288 & 0.398/0.375 & 0.402/0.391 \\
         CS & \textbf{0.834}/\textbf{0.849} & 0.477/0.513& 0.542/0.578& 0.703/0.724 &0.415/0.407 &0.321/0.297 &0.286/0.278 &0.301/0.282 \\
        CS + RaidaR & 0.714/0.791 & 0.547/0.589 & 0.582/0.663 & 0.712/0.734 
        & 0.421/0.409 & 0.361/0.337 & 0.354/0.332 & 0.373/0.319  \\
         CS + RaidaR + Syn & 0.769/0.813 & 0.593/0.631 & 0.625/0.672 & \textbf{0.719}/\textbf{0.741}
         & \textbf{0.434}/\textbf{0.412} & 0.369/0.340 & 0.366/0.348 & 0.385/0.351 \\
         BDD & 0.538/0.571 & 0.596/0.621 & 0.651/0.672& 0.622/0.673 
         & 0.334/0.315 & 0.371/0.362 & 0.386/0.370 & 0.362/0.338\\
         BDD + RaidaR & 0.603/0.618 & 0.638/0.652 & 0.676/0.706 & 0.681/0.725 
         & 0.351/0.338 & 0.384/0.369 & 0.395/0.378 & 0.392/0.384 \\
         BDD + RaidaR + Syn &0.612/0.615 & \textbf{0.644}/\textbf{0.654} & \textbf{0.678}/\textbf{0.714} & \textbf{0.719}/0.731 
         & 0.366/0.323 & \textbf{0.397}/\textbf{0.370} & \textbf{0.400}/\textbf{0.387} & \textbf{0.408}/\textbf{0.393}\\
    \end{tabular}
        \vspace{-2pt}
    \caption{Left: mIoU (averaged by class) for semantic segmentation using Vpred \cite{vpred} and HMSA \cite{hmsa} on cross-dataset training and testing. Right: AP for instance segmentation using PointRend \cite{pointrend} and TensorMask \cite{tensormask}. CS stands for Cityscapes.}
    \label{tab:cross-experiment-all}
     \vspace{-4pt}
\end{table*}

\vspace{4pt} 
\noindent\textbf{Statistics.}
Our 5,000 ground truth rainy images are composed of 1,738 images with droplets, 76 foggy, and 1,214 captured at night that are semantically segmented by 19 labels following the naming convention of Cityscapes. 
This shows the diversity of RaidaR for different raining scenarios.
Overall, the semantic masks in RaidaR include 37,126 traffic signs, 25,894 cars, 15,162 vegetation, 12,387 traffic lights, 9,175 sidewalks, 5,638 roads, 2,015 pedestrians, etc.
In addition, there are 84,602 and 3,977 instance masks for vehicle and pedestrian, respectively.

%% file: MainExperiments.tex


\section{Experiments}
\label{sec:experiment}
In this section, we conduct both quantitative and qualitative experiments to show how challenging our RaidaR dataset is for the visual perception tasks and its value as a \textit{complement} to existing datasets.
First, we compare the semantic and instance segmentation results on various cross-dataset training and testing configurations.
Then, we verify the effectiveness and efficiency of our cross-validation based label generation scheme for generating the ground truth semantic labels.
We also apply a de-raining method on our dataset and demonstrate that natural raining artifacts are not easy to remove and that training a segmentation network using such data is more effective than using a current de-raining method to obtain proper segmentation.




\subsection{Cross-dataset Training and Testing}
\label{sec:cross-testing}
\noindent\textbf{Semantic Segmentation.}
 The rainy images in our dataset can raise significant challenges for existing computer vision algorithms.
 To better position our dataset among the existing ones, e.g., for the semantic segmentation task,
 we compare the results of Vpred segmentation model \cite{vpred} in multiple cross-dataset experiments.
 Moreover, we conduct a separate experiment using HMSA (Hierarchical Multi-Scale Attention) \cite{hmsa}, the SOTA open source semantic segmentation model on the Cityscapes Benchmark.
 As HMSA is not used in our cross-validation based ground truth generation step, comparing the results from HMSA provides a more fair comparison about how well the SOTA segmentation algorithms perform on the experimented datasets.
 
 The standard class-averaged metric Intersection-over-Union (mIoU) values on Cityscapes (CS) \cite{cityscapes}, BDD \cite{BDD}, and our RaidaR are reported in Table \ref{tab:cross-experiment-all} (left) based on various cross-dataset training and testing configurations.
 Note that for RaidaR, we use the 5,000 annotated rainy images and split them by 7:1:2 for training, validation and testing.
 To verify whether our dataset can indeed complement the existing dataset such as BDD which also contains rainy images, we trained new models on the combined training dataset (BDD + RaidaR).
 Moreover, we combined BDD + RaidaR with a new synthetic dataset which contains 500 synthetic sunny and 500 synthetic rainy images generated from RaidaR by our masked image-to-image translation (Section \ref{sec:masked}) to see whether the performance can be further boosted.
 For each model, we also test their performance on the 253 rainy images extracted from BDD (i.e., BDD rainy).

 From Table \ref{tab:cross-experiment-all}, we can see the models trained on original non-combined datasets generally perform well on their own testing dataset.
 For cross-testing, e.g., the model trained on CS and tested on BDD, the mIoUs drop significantly which may be caused by the domain shifts between the datasets.
Similarly, the model trained on RaidaR alone does not perform well on CS or BDD, which are composed of \wl{almost exclusively} sunny images.
 On the other hand, the CS-trained model performs better on RaidaR than BDD \rui{since BDD is more diverse and difficult}.
When combining RaidaR with other datasets such as BDD, performance can be generally improved \rui{since RaidaR can complement the existing datasets for better performance on rainy images}.
\rui{Furthermore, the results on BDD rainy verify RaidaR's \wl{positive} contribution on rainy images.
\rui{One exception is the CS-trained model achieves better performance on CS comparing to the models trained on CS + RaidaR etc.
This is also reasonable since CS only contains sunny images while adding RaidaR may not help.
}
}

\begin{figure*}[t]
  \centering
  \includegraphics[width=0.95\linewidth]{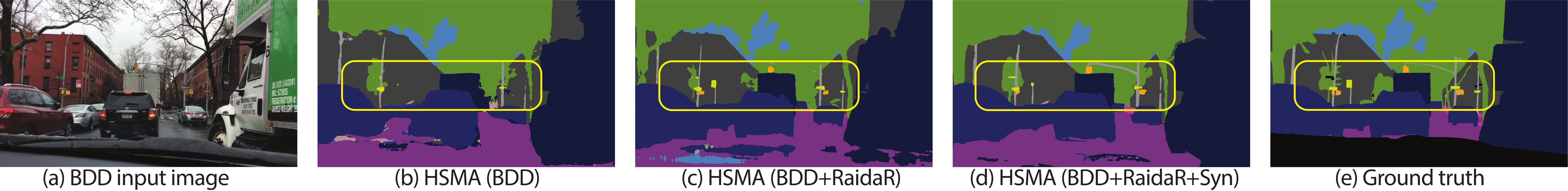}
    \vspace{-6pt}
  \caption{Qualitative comparisons of HMSA trained on different training configurations and tested on BDD. Highlighted regions by the yellow boxes show that after adding RaidaR or RaidaR + Syn, the model can produce finer details (e.g., traffic signs, trees). }
  \label{fig:segmentation_bdd_hmsa}
  \vspace{-4pt}
\end{figure*}

\begin{figure*}[t]
  \centering
  \includegraphics[width=0.95\linewidth]{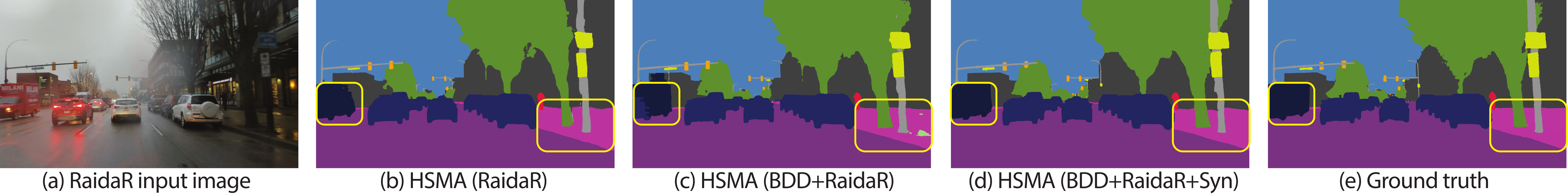}
    \vspace{-6pt}
  \caption{Qualitative comparisons of HMSA trained on different training configurations and tested on RaidaR. Highlighted regions by the yellow boxes show that the model can produce satisfactory results on rainy images when trained on RaidaR, but inferior results when simply combined with BDD. After adding the synthetic images (Syn), the model can produce improved results.}
  \label{fig:segmentation_ours_hmsa}
  \vspace{-6pt}
\end{figure*}

Finally, after adding synthetic images into BDD + RaidaR, the performance is generally improved.
 In fact, our masked image-to-image translation can be considered as an effective data augmentation to enhance the existing datasets.
 Figure \ref{fig:segmentation_bdd_hmsa} and \ref{fig:segmentation_ours_hmsa} verify the above observations about the combined datasets with results obtained using HMSA. 



 \noindent\textbf{Instance Segmentation.}
 To further verify the usefulness of RaidaR and our synthetic data, we also conduct instance segmentation experiments following the similar cross-dataset training and testing configurations for semantic segmentation.
 Specifically, we perform PointRend \cite{pointrend} and TensorMask \cite{tensormask}, two SOTA instance segmentation algorithms implemented in Detectron2 \cite{wu2019detectron2},
 and calculate Average Precision (AP) on vehicles and pedestrians.
From Table \ref{tab:cross-experiment-all} (right),
similar improvements of the AP values from PointRend and TensorMask can be observed when RaidaR and synthetic images are added to the training data.




\vspace{4pt} 
\noindent\textbf{Training Details.}
All of our models have been trained using the open source code available online.
For semantic segmentation, we used available models of Vpred \cite{vpred} and HMSA \cite{hmsa} which are pre-trained on Cityscapes, and continued training them separately on different datasets, i.e., RaidaR, CS + RaidaR, CS + RaidaR+ Syn, BDD, BDD + RaidaR, BDD + RaidaR + Syn.
For instance segmentation, we used available code and pre-trained models on Detectron2 \cite{wu2019detectron2} for PointRend \cite{pointrend} and TensorMask \cite{tensormask}, and fine-tuned the networks on different datasets.
Specifically, for Cityscapes, we directly used the pre-trained models for both semantic and instance segmentation.
For BDD, we used their train/val/test split with 7,000, 1,000,
2,000 images and trained the models for another about 20 epochs.
RaidaR (3500 out of 5000 rainy images for training) or Syn (500 synthetic sunny and 500 synthetic rainy images) are combined with CS or BDD and used to fine-tune the models for about 20 epochs.  

 Overall, the quantitative and qualitative results confirm our dataset can contribute to existing datasets for increasing their semantic and instance segmentation performance on rainy images.
 Note that a simple combination of some datasets may not be the optimal way to use their full strength.
 We leave more sophisticated explorations of how our dataset can further contribute to or benefit from other datasets to future work.

\begin{table}[t]
    \centering
    \small
    \begin{tabular}{c|c|c}
         \diagbox{Train}{Test}& RaidaR (De-rain) & RaidaR \\\hline
         CS & 0.705 & 0.701\\
         RaidaR (De-rain) & 0.712 & 0.707 \\
         RaidaR & 0.713& \textbf{0.716}   \\
    \end{tabular}
    \vspace{-4pt}
    \caption{mIoU (averaged by class) for Vpred trained and tested on Cityscapes (CS), de-rained RaidaR, and RaidaR.}
    \label{tab:de-rain}
    \vspace{-6pt}
\end{table}
\normalsize

\subsection{Effects of Rain Artifacts for Segmentation.}

To evaluate how raining artifacts complicate perception tasks (e.g., segmentation), we train Vpred on RaidaR and its de-rained version. To de-rain RaidaR, we utilize MPRNet \cite{MPR_net} on both training and testing sets of RaidaR. We then train Vpred on Cityscape, RaidaR (De-rain) and RaidaR. 
From Table~\ref{tab:de-rain}, we can observe that training on Cityscape and testing on RaidaR produces the worst outcome while training on RaidaR and testing on RaidaR produces the best one. This demonstrates that simply de-raining images is not effective to obtain good segmentation results and our RaidaR helps the network learn a more robust segmentation model over images with raining artifacts.

\subsection{Cross-Validation based Label Generation}
\label{sec:Cross_Validation}
\noindent\textbf{Effectiveness.} Our cross-validation based ground truth generation takes the advantages of existing segmentation algorithms to accelerate the image labeling process.
To verify its effectiveness, we quantitatively evaluated the percentage of pixels labeled as reliable by the automatic cross validation step with different weights $w_k$ (Equation \ref{eq:conf_score}) on a selection of 2,000 images.
We found that using 0.4, 0.3, 0.2, 0.1 for Vpred, Accel, PSP and PSA can generate more reliable labels than other weight combinations, i.e., 75.2\% of the pixels in an image comparing to 57.1\% which uses equal weights for each method.

\begin{figure}[t]
  \centering
  \includegraphics[width=0.95\linewidth]{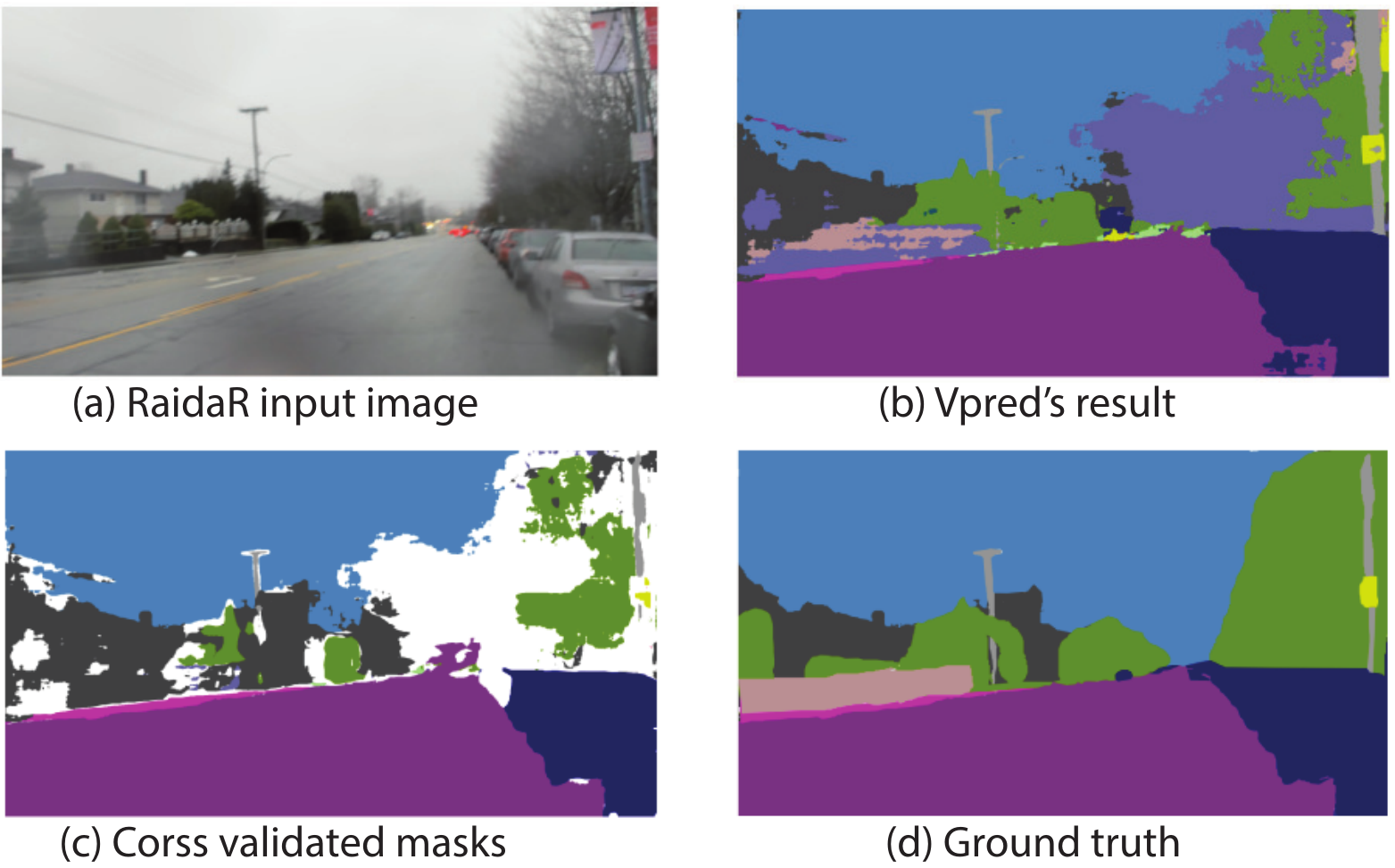}
  \vspace{-2pt}
  \caption{Having an input image (a), Vpred alone produces a segmentation (b) that needs improvements. The labels are cross validated with three other segmentation networks (c) where the inconsistent labels are identified by white pixels. Manual intervention is performed to determine the labels of these pixels and slightly tweak the other ones to obtain the ground truth (d).}
  \label{fig:cross-validation}
  \vspace{-4pt}
\end{figure}

Note that finding the optimal weight setting is not our goal.
Instead, we mainly use cross validation to reduce the manual effort of annotating the large number of easy-to-predict pixels.
Using the proposed weight setting, only 24.8\% of pixels are left to be manually annotated.
In Figure \ref{fig:cross-validation}, we show segmentation results of only using Vpred, the cross-validation approach and the final ground truth label after manual intervention.
It can be observed that the cross-validation approach can predict a reliable segmentation for most parts of the images, leaving some difficult or error-prone pixels for manual intervention.

To further verify that our cross-validation based approach can generate ground truth similar to the full manual annotation, we randomly select 200 images from Cityscapes and compare our generated ground truth with theirs. On average, less than 2.1\% of pixels in our ground truth are different from the manually annotated ground truth from Cityscapes. Note that the comparison didn't include the ``void'' label which relates to the windshield or hood of the ego-vehicle as our images are captured using a top-mounted camera and don't have such regions.

\vspace{4pt} 
\noindent\textbf{Efficiency.} To quantitatively evaluate how our cross-validation method increases the efficiency of the ground truth label annotation, we conduct a preliminary experiment to compare the labeling time of our process with the time of annotating from scratch. We asked five users to label 10 images on all original pixels from scratch. We then let those users label the same images again using our cross-validation based labeling process. After calculating the time spent on each image, on average, it takes about 8.6 minutes for labeling each image using our labeling process and 53.8 minutes by manually labeling all pixels using the same tool.

\begin{figure}[t]
  \centering
  \includegraphics[width=0.95\linewidth]{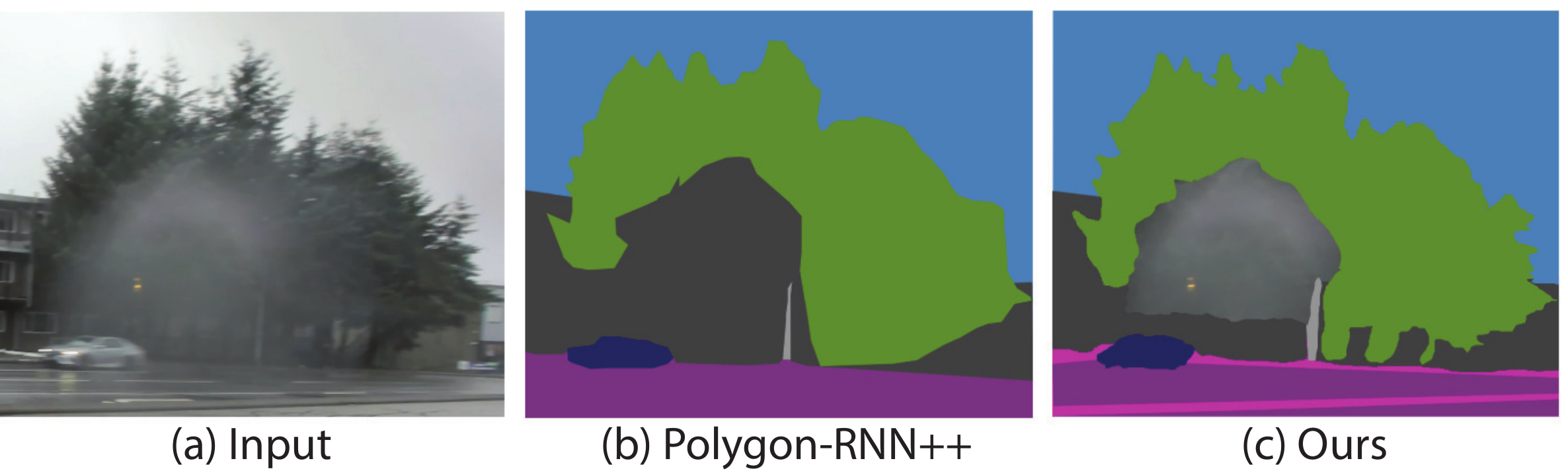}
  \vspace{-2pt}
  \caption{Given an input rainy image (a),  after initializing bounding boxes, Polygon-RNN++ generates (b). As apparent, our mechanism (automatically) generates better details around the droplet boundary in (c). Note that we keep the droplet region in (c) to indicate that this region needs manual intervention. No post-processing is applied on either mask.}
  \label{fig:comparison}
  \vspace{-4pt}
\end{figure}

\vspace{4pt} 
\noindent\textbf{Comparison with Polygon-RNN++.}
We compare our ground truth generation mechanism with Polygon-RNN++ \cite{acuna2018} which is a SOTA semi-automatic annotation technique (Curve-GCN's code \cite{ling2019} was not available at the time of this submission). In Polygon-RNN++, users are required to determine a bounding box to generate a mask for each object (Figure \ref{fig:comparison} (b)). Ours instead directly generates masks for all classes without manual intervention for each single class (Figure \ref{fig:comparison} (c)). In addition, Polygon-RNN++ does not fully reproduce thin parts or fine details around the boundary of objects, especially when the image contains weathering artifacts (e.g., fog, droplets). Therefore, it still needs considerable manual adjustments. In contrast, ours is based on current state-of-art segmentation algorithms which can accurately detect edges and details. Based on a small experiment on 10 images, generating a ground truth segmentation takes about 25 minutes per image using Polygon-RNN++ while our mechanism only needs 9 minutes.

%% file: MainMasked.tex
\section{Masked Image-to-Image Translation}
\label{sec:masked}



While applications such as training segmentation or detection networks for autonomous driving systems can directly benefit from our dataset, there are approaches that use semantic segmentation as prior to better perform a task, such as image-to-image translation \cite{cherian2019,roy2019}.
Here, we benefit from RaidaR segmentation for \textit{masked image-to-image translation} to convert rainy images to sunny and vice-versa.
This can be useful for improving the quality of images with weathering artifacts or data augmentation for training visual perception algorithms (Section \ref{sec:experiment}).

\begin{figure}[t]
  \centering
  \includegraphics[width=0.95\linewidth]{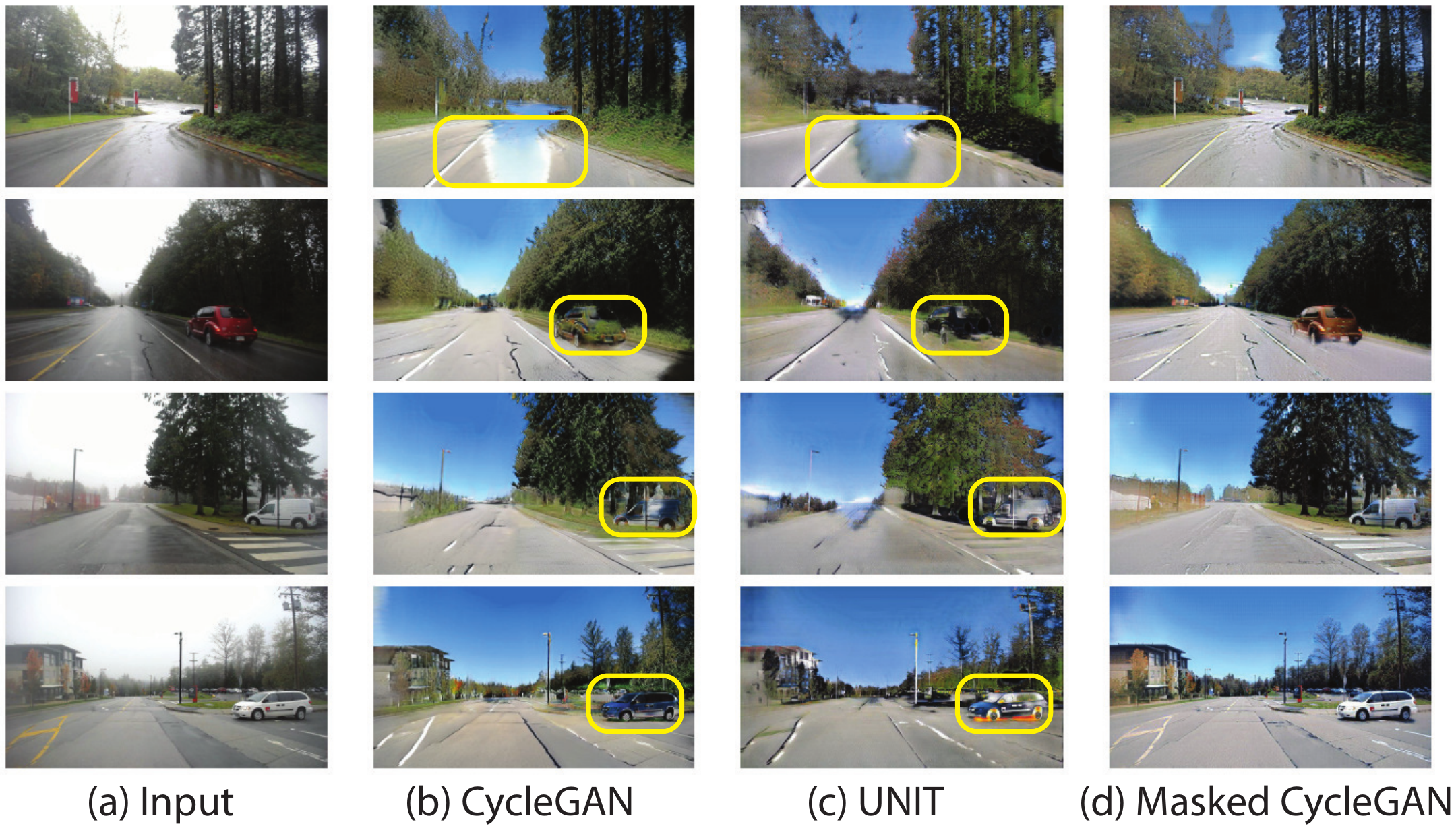}
  \caption{Inputs (a) and the results of CycleGAN (b), Unit (c), and our masked CycleGAN (d). As apparent, our approach is more successful in preserving the colors of the input image. The colors of some regions (highlighted by yellow box) are not preserved in CycleGAN and UNIT due to the lack of semantic awareness.}
  \label{fig:res1}
  \vspace{-4pt}
\end{figure}

\begin{figure}[t]
  \centering
  \includegraphics[width=0.95\linewidth]{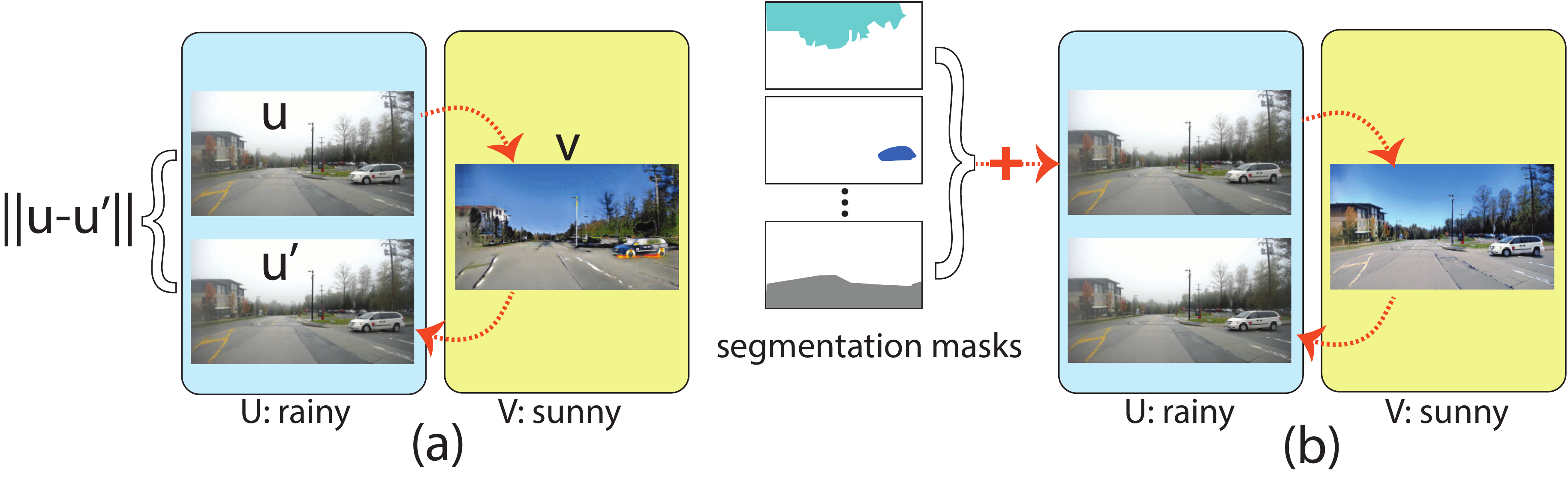}
  \caption{In CycleGAN, an image is translated from domain U (rainy) to V (sunny) (a). By returning to domain U in a cycle, the same images should be produced through a reconstruction loss. To instruct the network about pixels' semantics, segmentation masks are provided as extra inputs (b).}
  \label{fig:CG}
  \vspace{-4pt}
\end{figure}

\begin{figure}[t]
  \centering
  \includegraphics[width=0.95\linewidth]{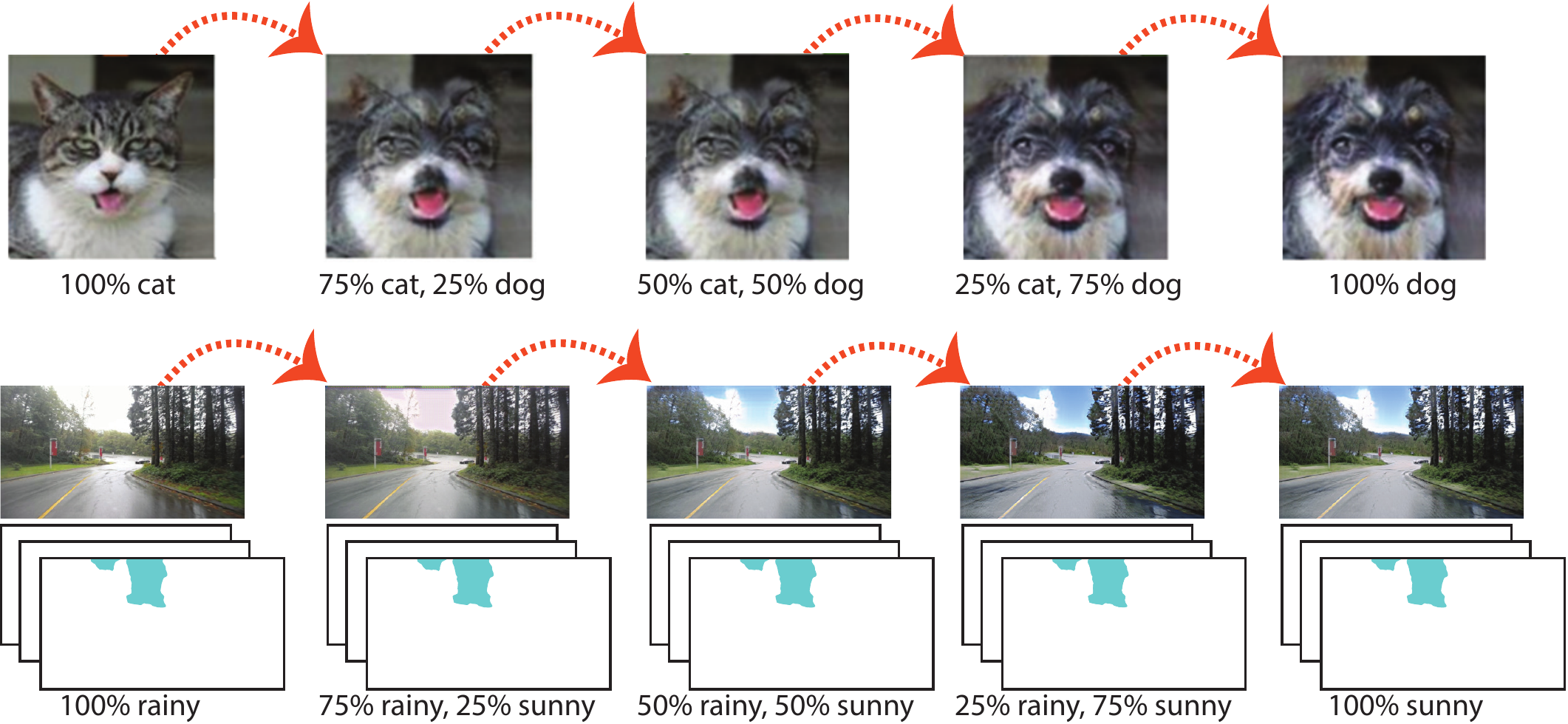}
  \caption{GANHopper translates an image from one domain to another in several hops (Top). We extend it to masked GANHopper by adding semantic masks for better transferring weathering artifacts (Bottom). }
  \label{fig:GH}
  \vspace{-4pt}
\end{figure}

Since it is impossible to capture the same road scene under two different weather conditions (e.g., sunny and rainy), we treat the problem as unpaired image translation. Among available methods, CycleGAN \cite{zhu2017} and UNIT \cite{liu2017} serve as strong baselines. However, both methods cannot transfer weathering artifacts successfully as they are only aware of the color distribution of pixels (Figure \ref{fig:res1} (b), (c)). 
Meanwhile, in a weathered road scene, two semantically different pixels (e.g., road and sky) may attain very similar colors (e.g., gray).
To resolve this problem, we introduce a masked CycleGAN in which semantic masks are fed to CycleGAN as additional inputs (Figure \ref{fig:CG}).
To account for semantic segments in CycleGAN, we modify its cycle loss by taking the masks into consideration and assigning higher weights to semantic regions whose colors should be preserved. 

\begin{figure}[t]
  \centering
  \includegraphics[width=0.95\linewidth]{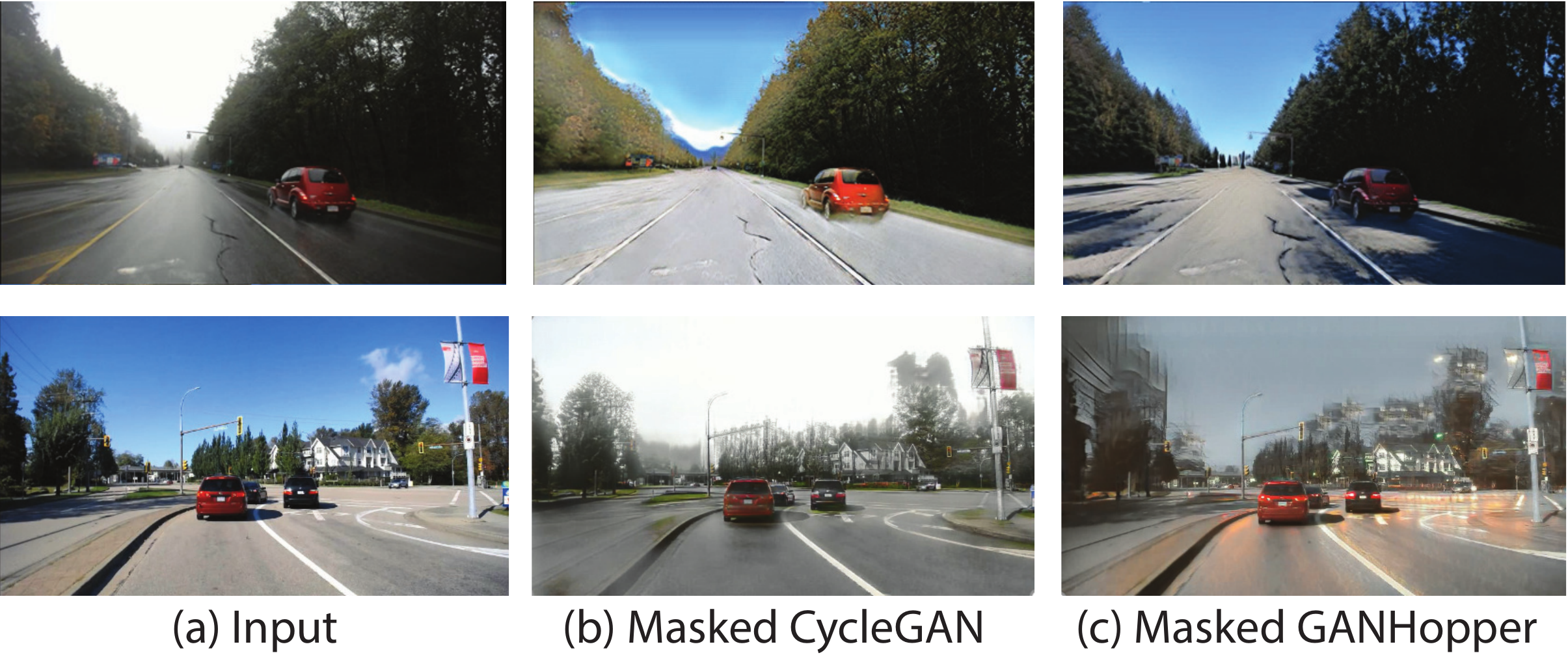}
  \caption{Both masked CycleGAN and masked GANHopper can produce plausible image translation. In comparison, masked GANHopper can better preserve the masked region and generate more delicate artifacts such as shadows in sunny scenes (Top) and reflections in rainy scenes (Bottom).}
  \label{fig:maskedGH}
  \vspace{-4pt}
\end{figure}

\vspace{4pt} 
\noindent\textbf{Masked GANHopper.}
To further improve the translation results, we extend the masked CycleGAN to GANHopper \cite{lira_eccv20} to benefit from a simple yet effective idea to perform the translation in multiple steps called \emph{hops} (Figure \ref{fig:GH}).
For masked GANHopper, we provide the masks to each hop and modify the loss function accordingly.
Benefited from RaidaR segmentation, both masked CycleGAN and masked GANHopper can produce visually plausible results for translating between sunny and rainy images.
Comparing to masked CycleGAN, masked GANHopper can perform better on preserving the masked region and produce relatively more delicate weathering artifacts as in Figure \ref{fig:maskedGH}.

\vspace{4pt} 
\noindent\textbf{Limitation and Discussion.}
\am{Note that in this section, we showed an application on how our RaidaR segmentation can be used as a prior for masked image-to-image translation, while not targeting at producing perfect image translation results nor most effective data augmentation for training.
Despite its plausible performance, using masked GANHopper may still produce some artifacts that cause difficulties for detecting far and fuzzy objects (e.g., a distant car with already very low confidence score).
On the other hand, by taking the masked region into account when image generation, the translated image can preserve the textures for pixels with important labels (e.g., traffic lights).
As more labels are passed as inputs, masked GANHopper will take slightly more inference time comparing to the non-masked version.
}

%% file: MainConclusion.tex
\section{Conclusion and Future Work}
\label{sec:conclusion}
We have provided a rainy rich dataset along with its semantic and instance segmentation ground truth.
We also presented a novel method to generate ground truth segmentation that can reduce the time and work-load needed for labeling pixels.
We showed that our dataset can be beneficial in training segmentation networks especially when the input is a rainy image. 
Our dataset can be advantageous in applications including training autonomous driving systems, droplet removal and image-to-image translation for which we provided a novel method that provides visually pleasing results that can be used to enhance the accuracy of segmentation networks. 
For future work, we intend to extend the variety and richness of our dataset in terms of road scenes (e.g., different cities), weathering artifacts (e.g., snow) and annotation types (e.g., lane markings).

\clearpage
\newpage

%% file: DataAnnotationSupp_arXiv_Appendix.tex
\appendix

\section{Appendix}
\label{sec:ref}

Here, we provide more details about our masked image-to-image translation frameworks discussed in the main paper, along with additional results of the semantic segmentation and instance segmentation.
A gallery of RaidaR image samples and their semantic and instance segmentation ground truth is also shown in Figure \ref{fig:gallery}.

\section{Masked Image-to-Image Translation Details}


In this section, we first describe the loss functions of our masked CycleGAN and masked GANHopper.
Then, we provide more implementation details and a preliminary comparison of inference time between the masked and the original image translation.

\subsection{Masked CycleGAN}

We have introduced a few adjustments to the original loss function of CycleGAN \cite{zhu2017} to respect semantic segments and control their influence on the final translation. Here, Equation~\ref{equation:masked_cyc_loss} describes the loss function for our masked CycleGAN.

\begin{equation}
\label{equation:masked_cyc_loss}
\begin{aligned}
    \Lagr_{cyc} = \sum_{i=1}^m \frac{\lambda_i}{p_i} |M_i \cdot F(G(u)) - M_i \cdot u|,
\end{aligned}
\end{equation}
In our notation, $u$ and $v$ respectively represent samples in domains $U$ and $V$ and $m$ is the number of labels in a segmentation mask.
$M_i$ is the binary mask for the $i$-th label, $p_i$ is the number of pixels with label $i$, and $\lambda_i$ is the weight for the $i$-th label. 
$\lambda_i$ controls the importance of each label and $p_i$ tries to give more influence to small but important regions. 
Note that $G$ and $F$ are the generators that translate images from domain $U$ to $V$ and vice-versa.
We have chosen $m=7$ labels in our segmentation masks: road, traffic lights, vegetation, sky, people, vehicles, and other. We set their respective $\lambda_i$ to $2,3,1,0.2,1,2,1$ in our model to distinguish the importance of different categories.

\subsection{Masked GANHopper}

We further modify masked CycleGAN to masked GANHopper, whose loss is shown in Equation~\ref{equation:mganh_loss} as:
\begin{equation}
\label{equation:mganh_loss}
\begin{aligned}
    \Lagr_{loss} =
    \gamma\Lagr_{\text{cyc}} +
    \epsilon\Lagr_{\text{adv}} +
    \delta\Lagr_{\text{dom}} + \zeta\Lagr_{\text{smooth}},
\end{aligned}
\end{equation}
same as the original GANHopper \cite{lira_eccv20}.  The cycle loss $\Lagr_{\text{cyc}}$ and the smoothness loss $\Lagr_{\text{smooth}}$ are both adapted to account for semantic segmentation masks provided in the dataset we propose in this paper, as shown in Equations~\ref{equation:mganh_cyc_loss} and \ref{equation:mganh_smooth_loss}. 
Let $h$ represent the number of hops and $G_n$ represent the transformation $G$ that is applied consecutively $n$ times, per GANHopper's framework.
The cycle loss and the smoothness loss in Equations~\ref{equation:mganh_cyc_loss} and \ref{equation:mganh_smooth_loss} that are defined for domain $U$ have analogous counterparts for domain $V$.
\begin{equation}
\label{equation:mganh_cyc_loss}
\begin{aligned}
    \Lagr_{cyc} = \sum_{n=1}^h \sum_{i=1}^m \frac{\lambda_i}{p_i} |M_i \cdot F(G_n(u)) - M_i \cdot G_{n-1}(u)|
\end{aligned}
\end{equation}
\begin{equation}
\label{equation:mganh_smooth_loss}
\begin{aligned}
    \Lagr_{smooth} = \sum_{n=1}^h \sum_{i=1}^m \frac{\lambda_i}{p_i} |M_i \cdot G_n(u) - M_i \cdot G_{n-1}(u)|
\end{aligned}
\end{equation}
The general purpose of these losses is the same as in the original CycleGAN and GANHopper. While the cycle loss aims to enforce the cycle consistency from one hop to the next, the smoothness loss aims to preserve the input image as much as possible.
These hop translations are represented as the functions $G$ and $F$ in this section.
These two losses have conflicting goals that tend to reach at an equilibrium as the network converges, which enables it to find intermediary domains to facilitate the translation process.

We trained our model with the same dataset configuration as masked CycleGAN but for 24 epochs. Since we optimize the parameters in each hop, GANHopper needs fewer epochs to train. We set $\gamma = 10, \epsilon=1, \delta =1, \zeta = 1$ in our model. All other variables have the same values as in masked CycleGAN.

\subsection{More Details and Inference Time Comparison}

For training masked CycleGAN and masked GANHopper, we used 20,791 rainy and 15,925 sunny images from RaidaR for 70 epochs.
Note that in this experiment, our goal is to train a generalizable model for the masked unpaired image-to-image translation.
Hence, we chose to use the larger set of RaidaR images with Vpred segmentation masks instead of our ground truth segmentation which is on a smaller set of RaidaR images (5,000 rainy and 4,085 sunny).
For testing and also generating the synthetic RaidaR dataset, we used the model trained above to a separate set of images with our ground truth segmentation masks.

We present a comparison of inference time for CycleGAN and masked CycleGAN in Table~\ref{tab:time_cmp}.
The inference times of GANHopper and masked GANHopper also demonstrate similar results since each hop of GANHopper uses the same architecture as CycleGAN. 
The results show that masked versions of these networks require a slightly longer processing time, but they are still fast and efficient.

\begin{table}[t]
    \centering
    \small
    \begin{tabular}{c|c|c}
          & Masked & Original \\ \hline
         Generator &4.9 - 5.9 ms& 3.8 - 4.8 ms\\ \hline
         Discriminator & 0.7 - 0.9 ms& 0.5 - 1.5 ms\\
    \end{tabular}
        \vspace{-4pt}
        \caption{Inference time for CycleGAN and masked CycleGAN. This table also represents inference for each hop in masked and original GANHopper. This is because the generator and discriminator from GANHopper have the same architecture as CycleGAN. }
    \label{tab:time_cmp}
    \vspace{-8pt}
\end{table}

\section{More Results}
\label{sec:dis}
\begin{figure*}
\centering
  \includegraphics[width=0.95\linewidth]{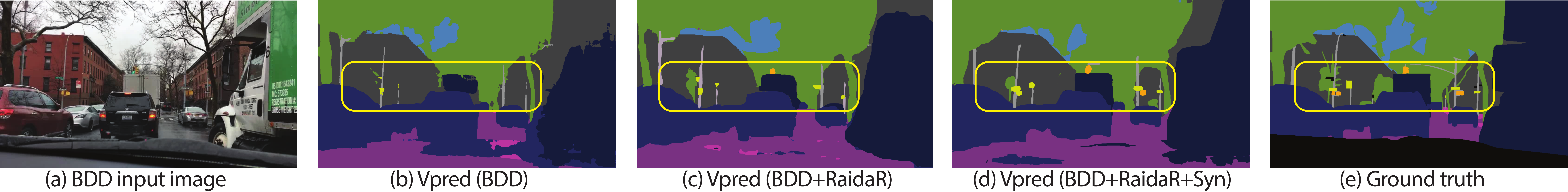}
          \vspace{-3pt}
  \captionof{figure}{Qualitative comparisons of Vpred trained on different training configurations and tested on BDD. Highlighted regions by the yellow boxes show that after adding RaidaR or RaidaR + Syn, the model is able to produce finer details (e.g., traffic signs, trees, etc).}
  \label{fig:segmentation_bdd}
\end{figure*}

\begin{figure*}
\centering
  \includegraphics[width=0.95\linewidth]{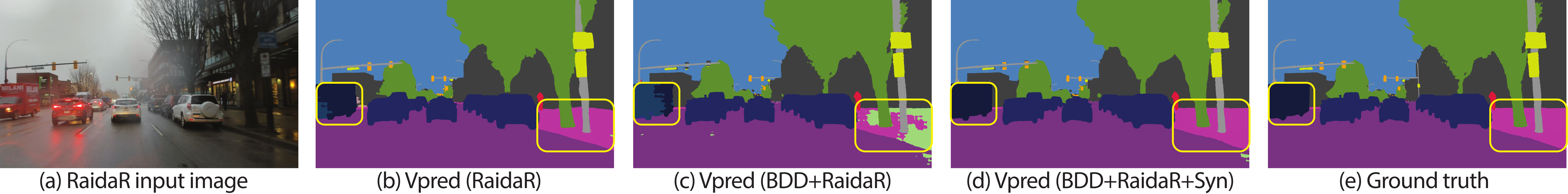}
        \vspace{-3pt}
  \captionof{figure}{Qualitative comparisons of Vpred trained on different training configurations and tested on RaidaR. Highlighted regions by the yellow boxes show that the model can produce satisfactory results on rainy images when trained on RaidaR, but inferior results when simply combined with BDD. After adding the synthetic images (Syn), the model can produce improved results.}
  \label{fig:segmentation_rr}
      \vspace{-8pt}
\end{figure*}

\begin{figure*}
\centering
  \includegraphics[width=0.95\linewidth]{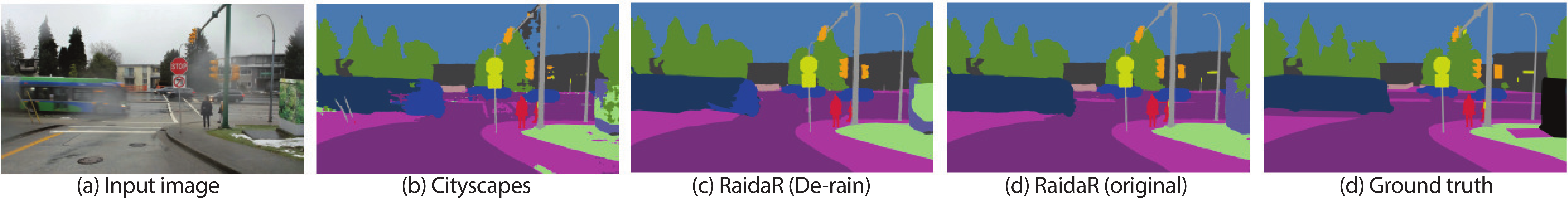}
    \vspace{-3pt}
  \captionof{figure}{Qualitative comparisons of Vpred trained on Cityscape, RadiaR (De-rain) and RaidaR (original).}
  \label{fig:de-rain}
\end{figure*}

\begin{figure*} [t]
\centering
  \includegraphics[width=0.95\linewidth]{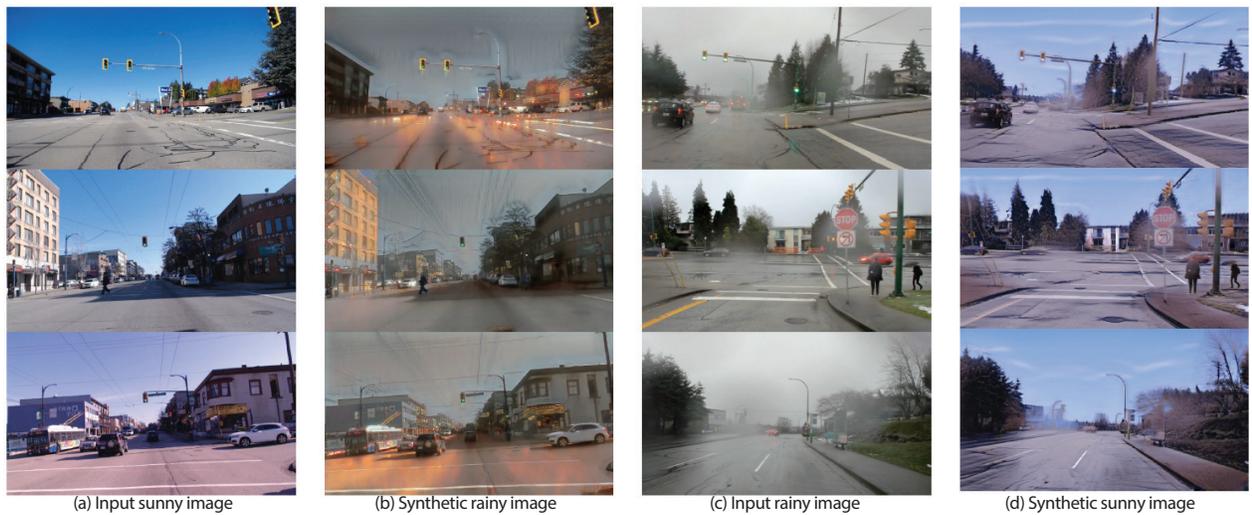}
    \vspace{-3pt}
  \captionof{figure}{More results of image translation using masked GANHopper trained on the RaidaR dataset.}
  \label{fig:ganh-synth}
\end{figure*}

\begin{figure*} [t]
\centering
  \includegraphics[width=0.95\linewidth]{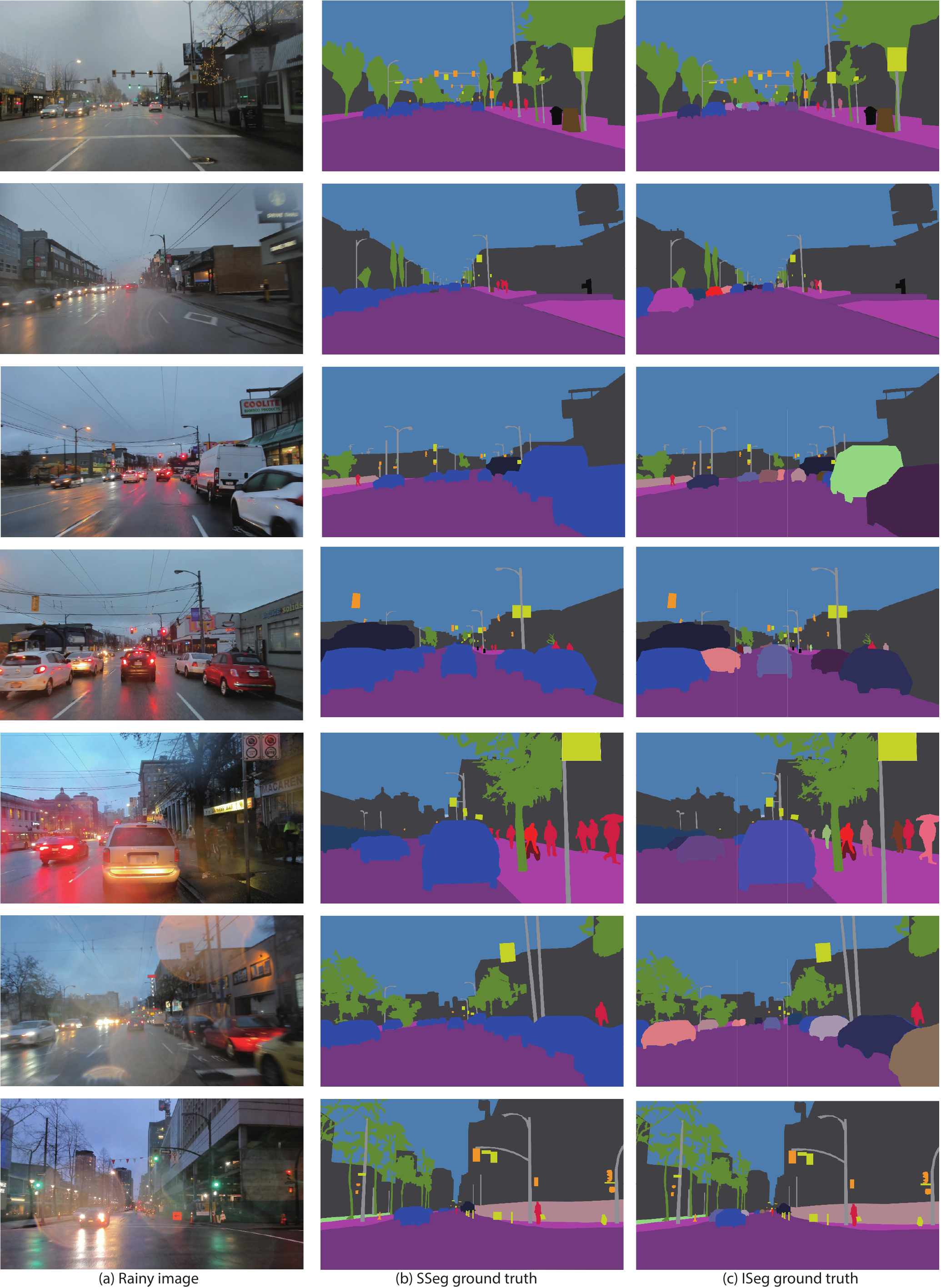}
    \vspace{-3pt}
  \captionof{figure}{Samples of RaidaR images and their semantic and instance ground truth masks.}
  \label{fig:gallery}
\end{figure*}

Here, we provide additional results for segmentation and masked image-to-image translation.
We also provide sample data of our rainy and sunny images along with their semantic and instance segmentation.
Figures \ref{fig:segmentation_bdd} and \ref{fig:segmentation_rr} show the results of Vpred \cite{vpred} trained on different training datasets and tested on BDD and RaidaR.
The results share similar observations as the corresponding experiment conducted using HMSA \cite{hmsa}. Detailed explanations of the results can be found in the main paper.
Figure \ref{fig:de-rain} shows the comparison with the de-raining based segmentation results.
It can be observed that the segmentation model trained directly on RaidaR outperforms the ones trained on Cityscapes or the de-rained version of RaidaR.
This verifies the usefulness of RaidaR for providing annotated rainy images to facilitate the segmentation tasks on images with rainy artifacts.
Figure~\ref{fig:ganh-synth} displays more RaidaR sample images and the corresponding results of image-to-image translation using masked GANHopper.
In the results, the overall style of images are successfully translated into the target domain while the colors of important categories (e.g., traffic light, cars) are well preserved.
Finally, Figure~\ref{fig:gallery} shows some samples of rainy images along with semantic and instance segmentation ground truth in RaidaR.

